\documentclass[runningheads,a4paper]{llncs}

\usepackage{hyphenat}
\usepackage{color}
\usepackage{amssymb}
\usepackage{hyperref}
\hypersetup{
    colorlinks=true,
    linkcolor=black,
    filecolor=black,      
    urlcolor=blue,
    citecolor=black,
}
\usepackage{graphicx}
\usepackage[misc,geometry]{ifsym} 
\usepackage{amsmath}
\DeclareMathOperator*{\argmax}{argmax}

\usepackage{url}
\begin{document}

\mainmatter  

\title{Muscle Excitation Estimation in Biomechanical Simulation Using NAF Reinforcement Learning}
\titlerunning{Deep RL-based Muscle Excitation Estimation}

\author{
Amir H. Abdi\textsuperscript{(\Letter)}
\and Pramit Saha
\and Venkata Praneeth Srungarapu
\and Sidney Fels
}

\authorrunning{Amir Abdi et al.}

\institute{Electrical and Computer Engineering Department, University of British Columbia, Vancouver, Canada\\
\email{amirabdi@ece.ubc.ca}}


\toctitle{RL-based Muscle Excitation Estimation}

\tocauthor{**********, *********, *************}
\maketitle


\begin{abstract}
Motor control is a set of time-varying muscle excitations which generate desired motions for a biomechanical system.
Muscle excitations cannot be directly measured from live subjects. 
An alternative approach is to estimate muscle activations using inverse motion-driven simulation.
In this article, we propose a deep reinforcement learning method to estimate the muscle excitations in simulated biomechanical systems.
Here, we introduce a custom-made reward function which incentivizes faster point-to-point tracking of target motion.
Moreover, we deploy two new techniques, namely, episode-based hard update and dual buffer experience replay, to avoid feedback training loops.
The proposed method is tested in four simulated 2D and 3D environments with 6 to 24 axial muscles.
The results show that the models were able to learn muscle excitations for given motions after nearly 100,000 simulated steps.
Moreover, the root mean square error in point-to-point reaching of the target across experiments was less than 1\%  of the length of the domain of motion.
Our reinforcement learning method is far from the conventional dynamic approaches as the muscle control is derived functionally by a set of distributed neurons. This can open paths for neural activity interpretation of this phenomenon.

\keywords{Deep reinforcement learning, Muscle excitation, Inverse motion-driven simulation, Normalized advantage function, Deep Q-network}
\end{abstract}


\section{Introduction}

Biomechanical modeling provides a powerful tool for analyzing structure and function of human anatomy, thereby establishing a scientific basis for
treatment planning.
Modeling is particularly indispensable in cases 
when mechanical variables are hard or impossible to measure with the available technologies~\cite{Pileicikiene2007}.
One of the unknown variables in understanding a musculoskeletal system is the muscle excitation trajectory. 
Muscle excitations reflect underlying neural control processes;
they form a connection between the causal neural activities and the resultant observed motion. Researchers have proposed various techniques, including forward-dynamics tracking simulation, inverse dynamics-based static optimization, and optimal control strategies, to predict muscle excitations and forces~\cite{erdemir2007model}.
However, the muscle redundancy problem
makes the task far more challenging. 
To address this problem, many approaches have been investigated including minimization of motion tracking errors, squared muscle forces and combined muscle stress~\cite{erdemir2007model}.
Unfortunately, the performance errors, high computational costs, and their sensitivity to optimal criteria and extensive regularizations have directed many researchers to seek alternative strategies~\cite{khan2017prediction,ravera2016estimation}.

Deep Reinforcement Learning (RL) is a popular area of machine learning that combines RL with deep neural networks to achieve higher levels of performance on  decision-making problems including  games, robotics and health-care~\cite{mnih2013playing}. 
In these approaches, an agent interacts with the environment and makes intelligent  decisions (actions) based on value functions $V(s)$, action-value functions $Q(s,a)$, policies $\pi(s,a)$, or  learned dynamics models.

Recent studies in deep RL have pushed the boundaries from discrete to continuous action spaces~\cite{Gu2016}, extending its possibilities for complex biomechanical control applications.
Although most researchers encode the locomotion in terms of joint angles for robotics applications, some progress has been made in the muscle-driven  RL-based  motion synthesis. 
Izawa \emph{et al.} introduced an actor-critic RL algorithm with subsequent prioritization of control action space to estimate the motor command of a biological arm model~\cite{izawa2004biological}.
Broad \emph{et al.} explored a receding horizon differential dynamic programming algorithm for arm dynamics optimization through muscle control policy to achieve desired trajectories 
in OpenSim~\cite{broad2011generating}.  These approaches require an \emph{a priori} information such as the relative action preferences or ranking systems 
which are rarely known in biomechanical systems. 
Quite recently, in the non-RL domain, 
deep learning approaches are also explored 
to predict  muscle excitations for point reaching movements, which rely on training data provided by inverse dynamics methods~\cite{Berniker2015,khan2017prediction}.

This work is aimed at developing an improved understanding of the effective coordination of muscle excitations to generate movements in musculoskeletal systems. 
It borrows ideas from and extends on a continuous variant of the deep Q-network (DQN) algorithm known as Normalized Advantage Function (NAF-DQN).
This research introduces three contributions: a \emph{customized reward function}, the \emph{episode-based hard update} of the  target model in double Q-learning, and the \emph{dual buffer experience replay}. It also takes advantage of a \emph{reduced-slope logistic function} to estimate muscle excitations.
The proposed approach is agnostic to the dynamics of the environment and is tested for systems with up to 24 degrees of freedom.
Due to its independence from training data, the trained models work as expected in unseen scenarios.

\section{Background}
The goal of reinforcement learning is to find a policy $\pi(a|s)$ which maximizes the expected sum of returns based on a reward function $r(s,a)$, where $a$ is the action taken in state $s$.
During  training, 
at each time step $t$,
the agent takes the action $a_t$
and arrives at state $s_{t+1}$ 
and is  rewarded with $R_t$, formulated as
\begin{equation}
R_t = \sum_{i=t}\gamma^{i-t}r(s_t, a_t),
\end{equation}
where $\gamma<1$ is a discount factor that reduces the value of future rewards.

In physical and biomechanical environments, system dynamics are not known and  a model of the environment, $p(s_{t+1}|s_t, a_t)$, cannot be directly  learned.
Therefore, a less direct model-free off-policy learning approach
is more beneficial.
Q-learning is an off-policy algorithm that learns a greedy deterministic policy based on the action-value function, $Q^\pi(s, a)$, referred to as the Q-function. Q-function determines how valuable  the $(s,a)$ tuple is under the policy $\pi$.
Q-learning is originally designed for  discrete action spaces and chooses the best action ($\mu$) as
\begin{equation}
\mu(s_t) = \argmax_a Q(s_t, a_t).
\end{equation}

Deep Q-network (DQN) is an extension of Q-learning that  uses a parameterized  value-action function, $\theta^Q$,  determined by a deep neural network. 
In DQN, the objective is to minimize the  Bellman error function
\begin{equation}
\label{eq:qlearning}
 \begin{array}{l}
L(\theta^Q) = \mathbb{E} [(Q(s_t,a_t|\theta^Q)-y_t)^2] \\
y_t=r(s_t,a_t) + \gamma Q(s_{t+1},\mu(s_{t+1})),
 \end{array}
\end{equation}
where $y_t$ is the observed discounted reward provided by the environment~\cite{Sutton1998}.

To enable DQN in continuous action spaces,
normalized advantage function (NAF) was introduced by Gu \emph{et al.}~\cite{Gu2016}, 
where the Q-function is represented as the sum of two parameterized functions, namely the value function, $V$, and the advantage function, $A$, as 
\begin{equation}
Q(s,a|\theta^Q) = V(s|\theta^V) + A(s,a).
\end{equation}
Value function describes a state and is simply the expected total future rewards of a state.
The Q function, on the other hand, describes an action in a state and explains how good it is to choose action $a$ at state $s$.
Therefore, the advantage function, defined as $Q(.)-V(.)$, is a notion of the relative importance of each action. 
NAF is defined as follows
\begin{equation}
\label{eq:advantage}
\begin{array}{l}
A(s,a) = -\dfrac{1}{2} (a-\mu(s|\theta^\mu))^TP(s|\theta^P)(a-\mu(s|\theta^\mu)) \\
P(s|\theta^P)=L(s|\theta^P)L(s|\theta^P)^T
\end{array},
\end{equation}
where $\mu$ is a parametric function determined by the neural network $\theta^\mu$; 
and $L$ is a lower-triangular matrix 
filled by the neural network $\theta^P$ with the diagonal terms squared; consequently, $P$ is a positive-definite square matrix~\cite{Gu2016}.
While there are numerous ways to define an advantage function,
the restricted parametric formulation of NAF ensures that the action that maximizes $Q^\pi$ is always given by $\mu(s|\theta^\mu)$.


\begin{figure}[t]
\centering
\includegraphics[width=0.98\textwidth, trim=1 1 1 1, clip]{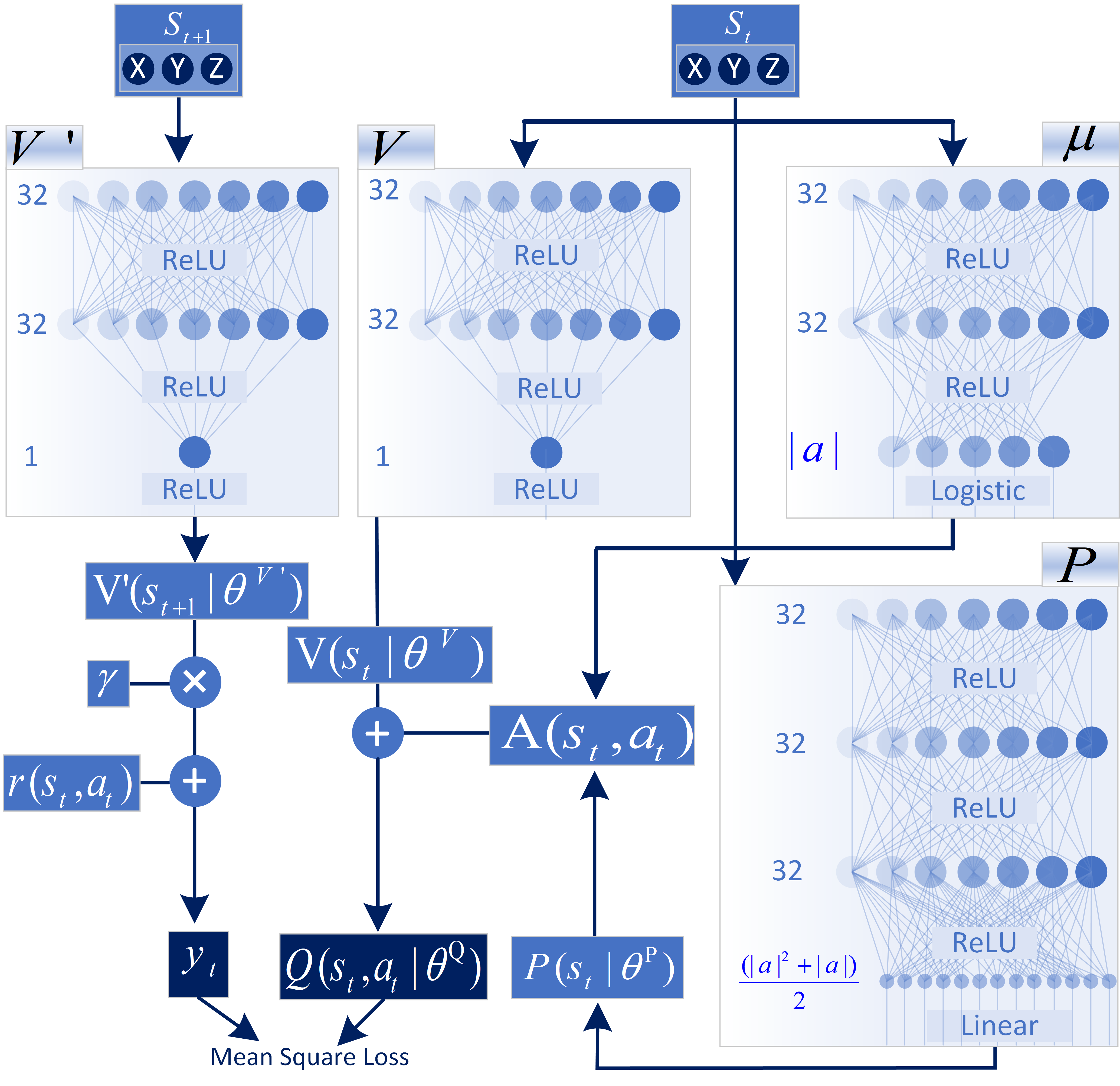}
\caption{The Normalized Advantage Function algorithm and architectures of the three neural networks, namely $\theta^\mu$, $\theta^V$, and $\theta^{L}$ (Eq.~3-5). The arrows demonstrate the data flow in the feedforward path.}
\label{fig:network}
\end{figure}

\section{Materials and Method}
\subsection{Model Architecture}
The deep dueling architecture used in this study is 
depicted in Fig.~\ref{fig:network}.
The $\theta^V$ and $\theta^{V'}$ neural networks receive the system state as input and estimate the value functions $V(s)$ and $V'(s)$.
This design follows the double Q-learning approach~\cite{van2016deep} 
to separate the action selector and evaluator operators.
During training, $\theta^{V'}$ receives the next state ($s_{t+1}$) to calculate the observed discounted total reward ($y_t$), while $\theta^V$ receives the current state ($s_t$) to predict the reward.

Here, we propose the \emph{episode-based hard update} technique, where
the  weights of the $\theta^{V'}$ network are only updated at the end of each episode by the weights of $\theta^V$.
This technique helps separate the exploration and training processes and mitigates the risk of getting stuck in a positive feedback loop.

The $\theta^P$ network receives the current state  and generates 
$(|a|^2+|a|)/2$ 
values to fill the lower triangular matrix of $L$, which, in turn, is squared to create matrix $P$ (Eq.~\ref{eq:advantage}). 
The $\theta^\mu$ network receives the current state, and estimates the  excitations of the $|a|$ muscles as a value between 0 and 1.
We deploy a \emph{reduced-slope logistic function}, $f(x)=1/(1+e^{-mx})$, 
in the final layer of the action selector $\theta^\mu$, 
where $0<m<1$ defines the slope.
This reduction in steepness mitigates the variance of muscle activations and results in a smoother muscle control.


\begin{figure}[t]
\centering
\includegraphics[width=1\textwidth, trim=5 5 0 5, clip]{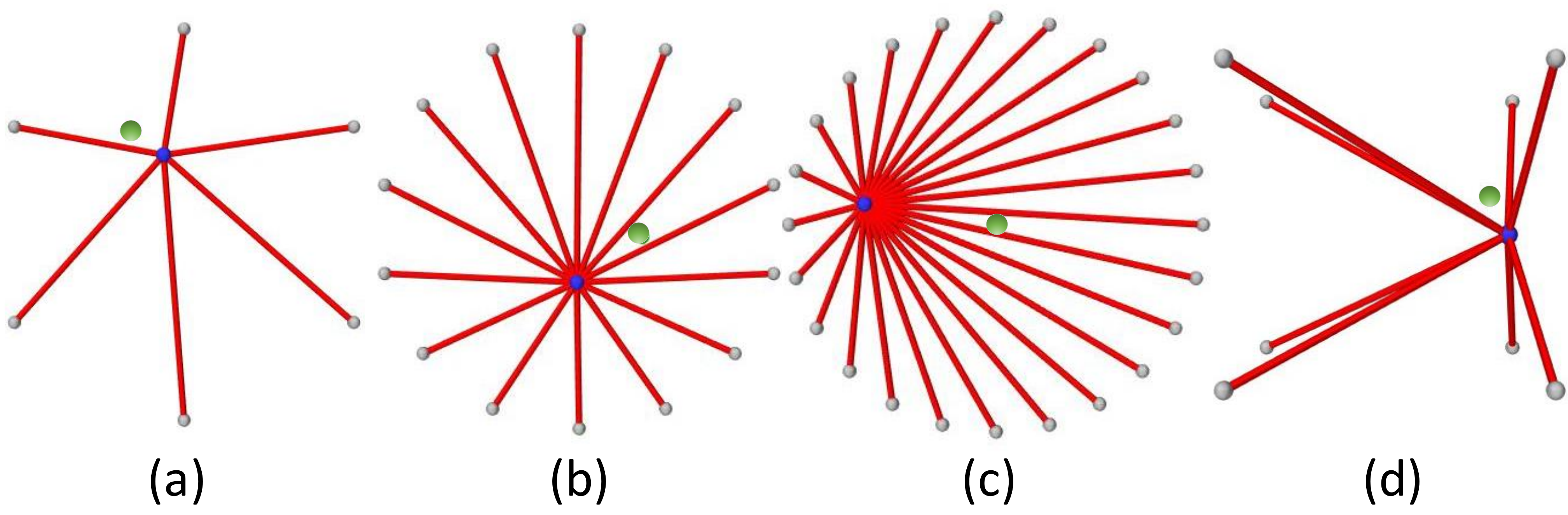}
\caption{The four simulation environments:   (a) 6-muscle 2D, (b) 14-muscle 2D, (c) 24-muscle 2D, (d) 8-muscle 3D. Notice the green and blue particles which visualize the position of the target ($T$) and the point mass ($P$).}
\label{fig:environments}
\end{figure}


\subsection{Methods} \label{training}

Training is composed of disjoint episodes. 
Each episode starts with a target point, $T$, randomly positioned in the motion space  of the point mass, $P$.
Motion space (domain) is defined as the entire area (volume) that the point mass can traverse in the simulated environment with muscle activations.
An episode ends either by the agent reaching the terminal state, \emph{i.e.} the point mass reaching the target, 
or going over a maximum number of steps per episode.

The size of the action space, $|a|$, is equal to the number of muscles. 
The state space is the spatial position of the target ($T_x, T_y, T_z$), where $T_y$ is constantly zero in the 2D simulations.
The position of the point mass (P) is  excluded from the state formulation
to imitate a real-world biomechanical system where the exact position of each joint is not known.

\subsubsection{Reward Function.} \label{reward}
The goal of the agent is to set the muscle excitations so that $P$ reaches $T$.
To incentivize this,
two factors were encoded in the reward function: distance and time,
\begin{equation}
\label{eq:reward}
r(s_t,a_t) =  \left\{
     \begin{array}{@{}l@{\thinspace}l}
       -1  \text{~~~~~} &    |P_{t+1}-T| \geq |P_{t}-T| \\
       1/t  &  |P_{t+1}-T| < |P_{t}-T| \\
       \omega &  |P_{t+1}-T| \leq d_{thres}\\      
     \end{array},
   \right.
\end{equation}
where $\omega>0$ is a constant value rewarded in a successful terminal state, and $|P~-~T|$ is the Euclidean distance between $P$ and $T$.
Terminal state is defined as a distance of less than $d_{thres}$ between $P$ and $T$.
The agent is penalized if $|P-T|$ is not lessened as a result of the action. 
Moreover, the rewards for correct decisions are reduced by a time factor to incentivize fast direct reach of the target, as opposed to curved trajectories.

\medskip
\noindent \textbf{Action Exploration.}
To enable action exploration in the continuous action domain, outputs of the $\theta^\mu$ network, \emph{i.e.} the muscle excitations, were augmented with a zero-mean stationary Gaussian-Markov stochastic process, known as the Ornstein Uhlenbeck.
The stochastic process was initialized with a variance of $0.35$, which was annealed as a function of $t$ down to $0.05$.

\medskip
\noindent \textbf{Dual Buffer Experience Replay.}
We borrow the idea of experience replay from the works of Mnih \emph{et al.} for higher data efficiency and training on non-consecutive samples~\cite{mnih2013playing}.
In this technique, a replay buffer stores  seen samples  as $(s_t,a_t,r_t,s_{t+1})$ tuples.
Samples are then then randomly selected for training from the buffer. 
In order to avoid feedback loops during an episode of training, we propose the \emph{dual buffer experience replay}
strategy,
where the \emph{episode buffer} stores samples of the current episode while the training samples are chosen from the \emph{back buffer}. 
At the end of each episode, contents of the \emph{episode buffer} are copied into the \emph{back buffer}.
This strategy is mostly important in the beginning of the training when the replay buffer is fairly sparse.

\medskip
\noindent \textbf{Training Hyper-parameters.}
If the success threshold radius, $d_{thres}$, is set too high, the agent is given the success reward too early and effortless; thus, will not learn the exact activation patterns. On the contrary, a small $d_{thres}$ value makes it impossible for the point mass to reach the success state which delays training.
Based on the above intuition, $d_{thres}$ was set so that  less than 1\% of the motion domain of the point mass is defined as the success state.
The learning rate ($\alpha$) and the reward discount factor ($\gamma$) were set to 0.01 and 0.99, respectively.
Each episode of training was constrained to 200 steps at the end of which the episode would restart and the state would reset to a new random position.


\section{Experiments and Results} \label{experiments}

\begin{figure}[t]
\centering
\includegraphics[width=\textwidth, trim=20 0 45 15, clip]{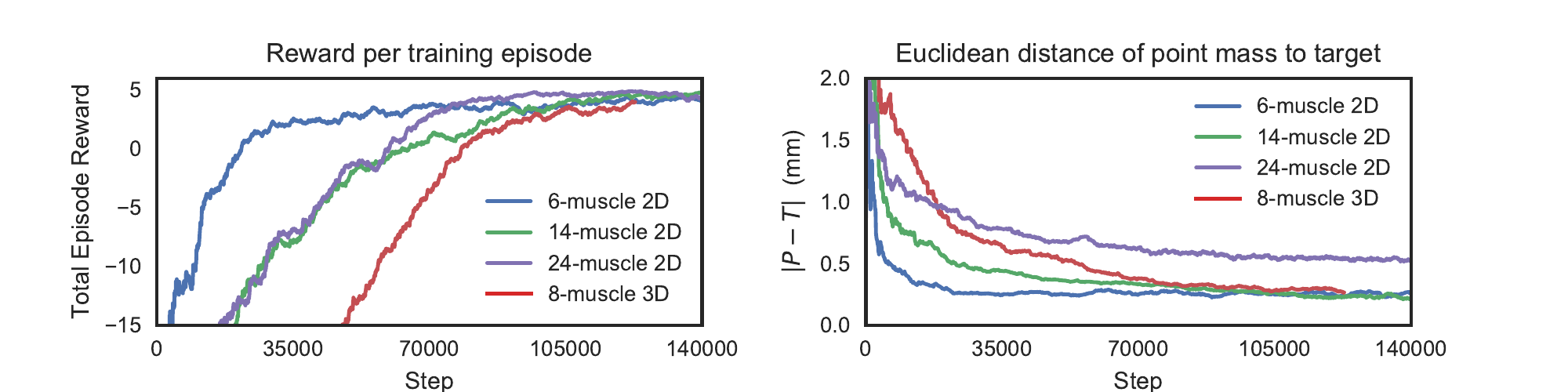}

\caption{Training of the RL agent in the four environments depicted in Fig.~\ref{fig:environments}.}
\label{fig:results}
\end{figure}

An open-source biomechanical simulator, ArtiSynth, was used to design the  simulation environments and run the experiments~\cite{Lloyd2012}. 
The keras-rl library with the TensorFlow backend was used as the basis for the implementations of the methods ~\cite{chollet2015keras,plappert2016kerasrl}.
A network interface was used to send the target positions from the simulated mechanical environment in ArtiSynth to the deep learning model and receive the new muscle activations from the model.
Our implementation, consisting of the ArtiSynth model in Java and the deep RL in Python, has been made available at
\href{https://github.com/amir-abdi/artisynth_point2point}{\url{https://github.com/amir-abdi/artisynth_point2point}}.
The forked keras-rl repository with added functionalities proposed in this paper can be accessed from
\href{https://github.com/amir-abdi/keras-rl}{\url{https://github.com/amir-abdi/keras-rl}}.

Four biomechanical environments were designed to test the feasibility and accuracy of the proposed method, including 2D environments with 6, 14, and 24 muscles, and a  3D environment with 8 muscles.
Muscles were set to have a maximum active isometric force  of $1$~N, optimal length  of $1$~cm, and maximum passive force of $0.1$~N, with $50\%$ flexibility for lengthening and shortening of fibers.
The damping coefficient was set to $0.1$.
The hyper-parameters of learning were not altered in between experiments for the results to be comparable.

In 2D environments, one end of each muscle was attached to the point mass,  while the other stationary end was positioned on the circumference of a circle of radius $10$~cm, as shown in Fig.~\ref{fig:environments}. 
In the 3D environment, the stationary ends of muscles were positioned at the 8 corners of a cuboid of length $20$~cm.
When the muscle excitations are zero, implying that the axial muscle fibers are at rest, the point mass will move to the very center of the circle. 
As the muscles get activated, 
the point mass moves towards the direction of the net force.
No other external force was applied to the point mass.

\subsection{Results}
Fig.~\ref{fig:results} demonstrates the weighted exponential smoothed curves for the total reward value  per episode and the distance between the point mass, $P$, and the target, $T$, at the end of each episode as  a function of the number of steps.
The agents were tested with 500 episodes of random point-to-point reaching tasks, and
the Euclidean distance between the final position of the point mass and the target was evaluated.
The root mean squared error (RMSE) of the trained agents were $1.8$~mm, $1.5$~mm, $1.7$~mm, and $2.4$~mm for the 6-muscle 2D, 14-muscle 2D, 24 muscle 2D, and 8-muscle 3D environments, respectively. The average RMSE across all environments was $1.8$~mm.

The learned models were also tested in unseen scenarios where the target point was moved out of the motion domain of the point mass. 
The surprising result was that the point mass followed the target, to the extent allowed by the model constraints, to minimize the distance between the two.

\section{Discussion and Conclusion}
In this article, a general reinforcement learning method was introduced to estimate muscle excitations for a given trajectory in a muscle-driven biomechanical simulation. 
The results assert that the approach is applicable to various degrees of freedom and muscle arrangements.
To make the method environment invariant, the agent was kept uninformed of the position of its associated point mass.
Moreover, the agent is unaware of the distribution of the muscles, their arrangement, their mechanical properties, and their states.
Therefore, it only receives  the location of the target point and the  rewards in response to its actions.

Deep reinforcement learning models are quite  sensitive to hyper-parameters and smaller neural networks have a higher chance of convergence. In our experiments, neural networks with more than 2 hidden layers for $\theta^\mu$ and $\theta^V$ did not converge.

As depicted in Fig.~\ref{fig:results}, the agent learned to reach the target in all environments irrespective of the number of muscles and their configurations. 
However, a positive correlation was observed between the training time and the degrees of freedom of the biomechanical system, \emph{i.e.}, the number of muscles.
The RMSE values indicate that trained agents managed to reach their target locations with a distance of less than half  the designated $d_{thres}$. 
In other words, the point mass has reached its target with less than 1\% distance with respect to its length of the domain of motion.
Interestingly, with no further fine tuning of the parameters, the performance of the method remained intact in higher degrees of freedom.

The results show that there exists a positive correlation between the $d_{thres}$ value and the final RMSE of the trained model.
However, setting  $d_{thres}$  to smaller values increases the chance of unsuccessful episodes which delays convergence.
Therefore, the authors suspect that gradually decaying this value, upon network convergence, can reduce the final RMSE of the model.

The proposed reinforcement learning method does not require any  labeled data for training, as opposed to other approaches where known optimal control trajectories were used as training data~\cite{Berniker2015,khan2017prediction}.
This highlights another important finding of the current study that
the trained models were functional when tested in unseen scenarios where the target point was moved out of the motion domain of the point mass. 
Since the model had learned the optimal muscle control independent of any training data, it was able to estimate the correct muscle excitations and move the point mass to minimize its distance to the target. 





The proposed reinforcement learning approach is different from the conventional inverse dynamics methods in the sense that the muscle controls are derived parametrically from  a set of distributed neurons of the  neural network, \emph{i.e}, $\theta^\mu$. 
Such approach opens the path for neural activity interpretation of the muscle control.


\bibliography{myBib} {}
\bibliographystyle{splncs03}

\end{document}